\documentclass[sn-mathphys]{sn-jnl}

\jyear{2022}%
\usepackage{multirow}
\usepackage[caption=false]{subfig}
\usepackage{makecell}
\usepackage{pifont}
\usepackage{xcolor}
\newcommand{\cmark}{\textcolor{green!80!black}{\ding{51}}}
\newcommand{\xmark}{\textcolor{red}{\ding{55}}}
\newcommand{\slfrac}[2]{\left.#1\middle/#2\right.}

\theoremstyle{thmstyleone}%
\theoremstyle{thmstyletwo}%

\theoremstyle{thmstylethree}%

\begin{document}
\title[Human-Centric Multimodal Machine Learning]{Human-Centric Multimodal Machine Learning: Recent Advances and Testbed on AI-based Recruitment}


\author{\fnm{Alejandro} \sur{Peña}}\email{alejandro.penna@uam.es}

\author{\fnm{Ignacio} \sur{Serna}}\email{ignacio.serna@uam.es}

\author{\fnm{Aythami} \sur{Morales}}\email{aythami.morales@uam.es}

\author{\fnm{Julian} \sur{Fierrez}}\email{julian.fierrez@uam.es}

\author{\fnm{Alfonso} \sur{Ortega}}\email{alfonso.ortega@uam.es}
\author{\fnm{Ainhoa} \sur{Herrarte}}\email{ainhoa.herrarte@uam.es}
\author{\fnm{Manuel} \sur{Alcantara}}\email{manuel.alcantara@uam.es}

\author{\fnm{Javier} \sur{Ortega-Garcia}}\email{javier.ortega@uam.es}

\affil{\orgname{Universidad Aut\'onoma de Madrid}, \orgaddress{\city{Madrid}, \postcode{28049}, \country{Spain}}}




\abstract{The presence of decision-making algorithms in society is rapidly increasing nowadays, while concerns about their transparency and the possibility of these algorithms becoming new sources of discrimination are arising. There is a certain consensus about the need to develop AI applications with a Human-Centric approach. Human-Centric Machine Learning needs to be developed based on four main requirements: (i) utility and social good; (ii) privacy and data ownership; (iii) transparency and accountability; and (iv) fairness in AI-driven decision-making processes. All these four Human-Centric requirements are closely related to each other. With the aim of studying how current multimodal algorithms based on heterogeneous sources of information are affected by sensitive elements and inner biases in the data, we propose a fictitious case study focused on automated recruitment: FairCVtest. We train automatic recruitment algorithms using a set of multimodal synthetic profiles including image, text, and structured data, which are consciously scored with gender and racial biases. FairCVtest shows the capacity of the Artificial Intelligence (AI) behind automatic recruitment tools built this way (a common practice in many other application scenarios beyond recruitment) to extract sensitive information from unstructured data and exploit it in combination to data biases in undesirable (unfair) ways. We present an overview of recent works developing techniques capable of removing sensitive information and biases from the decision-making process of deep learning architectures, as well as commonly used databases for fairness research in AI. We demonstrate how learning approaches developed to guarantee privacy in latent spaces can lead to unbiased and fair automatic decision-making process. Our methodology and results show how to generate fairer AI-based tools in general, and in particular fairer automated recruitment systems.\footnote{Paper based on the keynote by Prof. Julian Fierrez at ICPRAM 2021.} }




\maketitle

\section{Introduction}\label{sec:Introduction}

Artificial Intelligence plays a key role in people's lives nowadays, with automatic systems being deployed in a large variety of fields, such as healthcare, education, or jurisprudence. The data science community's breakthroughs of the last decades along with the large amounts of data currently available have made possible such deployment, allowing us to train deep models that achieve a performance never seen before. The emergence of deep learning technologies has generated a paradigm shift, with handcrafted algorithms being replaced by data-driven approaches. However, the application of machine learning algorithms built using training data collected from society can lead to adverse effects, as these data may reflect current socio-cultural and historical biases~\cite{BigDataImpact}. In this scenario, automated decision-making models have the capacity to replicate human biases present in the data, or even amplify them~\cite{acien2018bias}\cite{demographic_bias_biometric}\cite{DL_prejudiced}\cite{MenAlsoLike}\cite{noble2018algorithms} if appropriate measures are not taken.

There are relevant models based on machine learning that have been shown to make decisions largely influenced by demographic attributes in various fields. For example, Google's \cite{Disc_Google} and Facebook's \cite{Disc_facebook} ad delivery systems generated undesirable discrimination with disparate performance across population groups. In $2016$, ProPublica researchers~\cite{Machine_bias} analyzed several Broward County defendants' criminal records $2$ years after being assessed with the recidivism system COMPAS, finding that the algorithm was biased towards black defendants. New York’s insurance regulator probed UnitedHealth Group over its use of an algorithm that researchers found to be racially biased, the algorithm prioritized healthier white patients over sicker black ones \cite{wallstreet2019bias}. Apple Credit service granted higher credit limits to men than women even though it was programmed to be blind to that variable~\cite{apple2019bias}. Face analysis technologies have also shown a gap in performance between some demographic groups~\cite{acien2018bias}\cite{Gender_shades}\cite{wang2020mitigating}\cite{serna2020formulation} as a major consequence of an undiverse representation of society in the training data. Moreover, as Balakrishnan \textit{et al.} pointed out~\cite{balakrishnan2020towards}, the problem of data bias goes beyond the training set, as we need a bias-free evaluation set in order to correctly assess algorithmic fairness.

The usage of AI technologies is also growing in the labor market~\cite{Hiring_Algorithms_Report}, where automatic-decision making systems are commonly used in different stages within the hiring pipeline~\cite{black2020ai}. However, automatic tools in this area have also exhibited worrying biased behaviors, such as Amazon's recruiting tool preferring male candidates over female ones~\cite{amazon2018bias}. Ensuring that all social groups have equal opportunities in the labor market is crucial to overcome differences with minority groups, which have been historically penalized~\cite{bertrand2004emily}. Some works are starting to address fairness in hiring~\cite{raghavan2020mitigating}\cite{schumann2020we}\cite{sanchez2020does}, but the lack of transparency (i.e. both the models and their training data are usually private for legal or corporate reasons~\cite{raghavan2020mitigating}) hinders the technical evaluation of these systems.

In response to the deployment of automatic systems, along with the concerns about their fairness, the governments are adopting regulations in this matter, placing special emphasis on personal data processing and preventing algorithmic discrimination. Among these regulations, the European Union's General Data Protection Regulation (GDPR)\footnote{https://gdpr.eu/} is specially relevant for its impact on the use of machine learning algorithms ~\cite{Right_explanation}. The GDPR aims to protect EU citizens' rights concerning data protection and privacy by regulating how to collect, store, and process personal data (e.g. Articles $17$ and $44$), and requires measures to prevent discriminatory effects while processing sensitive data (according to Article $9$, sensitive data includes ``personal data revealing racial or ethnic origin, political opinions, religious or philosophical beliefs"). Thus, research on transparency, fairness, or explicability in machine learning is not only an ethical matter, but a legal concern and the basis for the development of responsible and helpful AI systems that can be trusted~\cite{cheng2021socially}. 

On the other hand, one of the most active areas in Machine Learning (ML) is around the development of new multimodal models capable of understanding and processing information from multiple heterogeneous sources of information \cite{Baltruaitis2017MultimodalML}. Among such sources of information we can include structured data (e.g. tabular data), and unstructured data from images, audio, and text. The implementation of these models in society must be accompanied by effective measures to prevent algorithms from becoming a source of discrimination. In this scenario, where multiple sources of both structured and unstructured data play a key role in algorithms' decisions, the task of detecting and preventing biases becomes even more relevant and difficult.

In this environment of desirable fair and trustworthy AI, the main contributions of this work are:
\begin{itemize}
\setlength\itemsep{0em}
    \item We review the latest advances in Human-Centric ML research with special focus on the public available databases proposed by the community. \item We present a new public experimental framework around automated recruitment, aimed to study how multimodal machine learning is influenced by demographic biases present in the training datasets: FairCVtest.\footnote{https://github.com/BiDAlab/FairCVtest}
    \item We have evaluated the capacity of both pretrained models and data-driven technologies to extract demographic information and learn biased target functions from multimodal sources of information, including images, texts, and structured data from resumes. 
    \item We evaluated a discrimination-aware learning method based on the elimination of sensitive information such as gender or ethnicity from the learning process of multimodal approaches, and apply it to our automatic recruitment testbed for improving fairness among demographic groups.
\end{itemize}

Our results demonstrate the high capacity of commonly used learning methods to expose sensitive information (e.g. gender and ethnicity) from different data domains,  and the necessity to implement appropriate techniques to guarantee discrimination-free decision-making processes. 

A preliminary version of this article was published in~\cite{FairCv}. This article significantly improves~\cite{FairCv} in the following aspects:
\begin{itemize}
\setlength\itemsep{0em}
    \item We extend FairCVdb to incorporate a name and a short biography to each profile. To the best of our knowledge, this upgrade makes FairCVdb the first fairness research database including image, text and structured data.
    \item We provide more extensive experiments within FairCVtest, where we analyze the impact of data bias on an automatic recruitment tool under different scenarios. In these experiments, we use commonly used fairness criteria to quantify this impact. We also measure the sensitive information exploited in the decision-making process, whereas~\cite{FairCv} limited the experiments to a more qualitative analysis. Furthermore, by including text data to our dataset, we extend FairCVtest with Natural Language Processing techniques.
    \item We provide a survey on fairness research in AI, in which we review some of the methods proposed in recent years to prevent algorithmic discrimination, and the most commonly used databases in the field.
\end{itemize}

The rest of the paper is structured as follows:
Section~\ref{sec:fairness_research} presents an overview on explainability in ML models, discrimination-aware ML approaches, and Human-Centric ML databases. Section~\ref{sec:FairCVdb} describes the considered automatic hiring pipeline, examines the information available in a resume highlighting the sensitive data associated to it, and describes the dataset created in this work: FairCvdb. Section~\ref{sec:problem_formulation} presents the general framework for our work including problem formulation. Section~\ref{sec:experiments} reports the experiments in our testbed FairCVtest after describing the experimental methodology and the different scenarios evaluated. Finally, Section~\ref{sec:conclusions} summarizes the main conclusions.

\section{Human-Centric Research in Machine Learning}
\label{sec:fairness_research}

The recent advances in AI and the large amounts of data available have made possible the deployment of automatic decision-making algorithms in our society. Due to their great impact in people's lives, especially in high stake settings, is essential that these systems are responsible and trustworthy. However, there are many models that have been shown to make decisions based on attributes considered as private (e.g. gender\footnote{We are aware of the studies that move away from the traditional view of gender as a binary variable~\cite{richards2016non}, and the difference between gender identity and biological sex. Despite the limitations of such model~\cite{keyes2018misgendering}, in this paper we use ``gender'' to refer to the external perception of biological sex, in line with the work historically developed in gender classification into male and female individuals.} and ethnicity), or exhibiting systematically discrimination against individuals belonging to disadvantaged groups. We can find examples of such unfair treatment in various fields, such as healthcare~\cite{wallstreet2019bias}\cite{larrazabal2020genderHC}, ad delivery systems~\cite{Disc_Google}\cite{Disc_facebook}\cite{speicher2018potentialAD}, hiring~\cite{Hiring_Algorithms_Report}\cite{amazon2018bias}, and both facial analysis~\cite{MenAlsoLike}\cite{Gender_shades}\cite{wang2020mitigating} and NLP technologies~\cite{Bias_in_Bios}\cite{bolukbasi2016man}. 

In the following sections we will present recent advances in Human-Centric ML research related with: 1) explainability and interpretability of ML models; 2) discrimination-aware ML approaches; and 3) databases for Human-Centric ML research.

\subsection{Interpretable and Explainable ML}

One of the long-term goals in deep learning is to learn abstract representations, which are generally invariant to local changes in the input \cite{bengio2013representation}. It has been observed that many learned representations correspond to human-interpretable concepts. But it is not quite clear what function they serve and whether it has a causal role that reveals how the network models its higher-level notions \cite{bau2020understanding}. Research is showing that not all representations in the convolutional layers of a DNN correspond to natural parts, raising the possibility of a different decomposition of the world than humans might expect, calling for further study into the exact nature of the learned representations \cite{yosinski2015visualization}\cite{geirhos2019bias}.

There is significant work on understanding neural networks. Most methods typically focus on what a network looks at when making a decision \cite{bach2015lrp}\cite{selvaraju2017grad}; other approaches seek to train explanatory models~\cite{ortega2021symbolic} or networks~\cite{hendricks2016explanations} that generate human-readable text.

We can distinguish between two types of approaches for generating a better understanding of an AI model: interpretable and explainable. As defined in \cite{montavon2018understanding}, an interpretation is the mapping of an abstract concept (e.g. a predicted class) into a domain that the human can make sense of, e.g. images or text; and an explanation is the collection of features of the interpretable domain that have contributed for a given example to produce a decision.

On the interpretation side we have Activation Maximization, which consists of looking for an input pattern that produces a maximum response of the model. It was introduced in \cite{erhan2009visualizing}, but such visualization technique has a limitation: as complexity increases, it becomes more difficult to find a simple representation of a higher layer unit, because the optimization does not converge to a single global minimum. Simonyan et al. came up with the suggestion to perform the optimization with respect to the input image, obtaining an artificial image representative of the class of interest \cite{simonyan2013visualizing}.

One way of improving activation maximization to enable enhanced visualizations of learned features is with the so-called expert. That is, in the function to be maximized, the L$2$-norm regularizer (a term that penalizes inputs that are farther away from the origin) is replaced by a more sophisticated one, called expert \cite{yosinski2015visualization}\cite{mahendran2015understand}\cite{nguyen2017visualize}\cite{nguyen2016visualization}. Another way is via deep generative models, incorporating such model in the activation maximization framework \cite{nguyen2016visualize}.

On the explanation side we have Sensitivity Analysis: how much do changes in each pixel affect the prediction. Initially intended for pruning neural networks and reducing the dimensionality of their input vector, was particularly useful for understanding the sensitivity of performance with respect to their structure, parameters, and input variables \cite{karnin1990sensitivity}\cite{zurada1994sensitivity}. More recently, it has been used for explaining the classification of images by deep neural networks. Simonyan et al. \cite{simonyan2013visualizing} applied partial derivatives to compute saliency maps. They show the sensitivity of each of the input image pixels, where the sensitivity of a pixel measures to what extent small changes in its value make the image to belong more or less to the class (local explanation).

Alternatives for explaining deep neural network predictions are backward propagation techniques. Some are: deconvolution, layer-wise relevance propagation (LRP) and guided backprop. 

Zeiler  and  Fergus \cite{zeiler2014visualizing} proposed deconvolution to compute a heatmap showing which input pattern originally caused a certain activation in the feature maps. The idea behind the deconvolution approach is to map the activations from the network back to pixel space using a backpropagation rule. The quantity being propagated can be filtered to retain only what passes through certain neurons or feature maps.

The LRP method \cite{bach2015lrp} applies a propagation rule that distributes back (without gradients) the classification output $f(x)$ decomposed into pixel relevances onto the input variables. This algorithm can be used to visualize the contribution of pixels both for and against a given class.

Guided backprop is the extension of the deconvolution approach for visualizing features learned by CNNs. Proposed in \cite{springenberg2015backprop}, it combines backpropagation and deconvolution by masking out the values for which at least one of the entries of the top gradient (deconvnet) or bottom data (backpropagation) is negative.

Another very well known backpropagation-based method combining gradients, network weights, and activations is Grad-CAM \cite{selvaraju2017grad}. Gradient-weighted Class Activation Mapping (Grad-CAM) uses the gradients of the class score with respect to the input image to produce a coarse localization map highlighting the important regions in the image for predicting the concept. It can be combined with guided backpropagation for fine-grained visualizations of  class-discriminative features.

These methods selectively illustrate one of the multiple patterns a filter represents, explanatory graphs provide a workaround. \cite{zhang2018interpreting} proposed a method disentangling part patterns from each filter to represent the semantic hierarchy hidden inside a CNN.

Some other methods have gone beyond visualization of CNNs and diagnosed CNN representations to gain a deep understanding of the features encoded in a CNN. Others report the inconsistency of some widely deployed saliency methods, as they are not independent of both the data on which the model was trained and the model parameters \cite{adebayo2018sanity}.

Szegedy et al. \cite{szegedy2014intriguing} reported the existence of blind spots and counter-intuitive properties of neural networks. They found that its possible to change the network's prediction by applying an imperceptible optimized perturbation to the input image, which they called and adversarial example. Paving the way for a series of works that sought to produce images with which to fool the models \cite{koh2017understanding}\cite{nguyen2015fool}\cite{su2019attack}.

Other studies aiming to understand deep neural networks are neuron ablation techniques. These seek a complete functional understanding of the model, trying to elucidate its inner workings or shed light on its internal representations. Bau et al. found evidence for the emergence of disentangled, human-interpretable units (of objects, materials and colors) during training \cite{bau2020understanding}.

\subsection{Discrimination-aware Learning}

In order to prevent automated systems from making decisions based on protected attributes or reproduce biased behaviors against disadvantaged groups, the research community has devised various ways to improve fairness in AI systems. These approaches are usually divided in the literature between pre-processing, in-processing, and post-processing techniques~\cite{cheng2021socially}.

The pre-processing techniques aim to transform the input domain to prevent discrimination and remove sensitive information. The authors of~\cite{quadrianto2019discovering} propose to remove sensitive information while improving model interpretability by learning a data-to-data transformation in the input domain, where the new representation achieves certain fairness criterion. This transformation is based in both neural style transfer and kernel Hilbert spaces. A similar approach is proposed in~\cite{FairnessGAN}, which seeks to generate a new dataset similar to a given one, but fairer with respect to a certain protected attribute. For this purpose, a fairness criterion is added to the loss function of an auxiliary GAN~\cite{AC-GAN}. In~\cite{calmon2017optimized} the authors address the pre-processing transformation as an optimization problem which trades off discrimination and utility at probabilistic level, while controlling sample distorsion on an individual level. More recently, Ramaswamy \textit{et al.} proposed~\cite{ramaswamy2021fair} a method for augmenting real datasets with GAN-generated synthetic images by modifying vectors in the GAN latent space to de-correlate sensitive and target attributes. 

In-processing approaches focus on the learning process as the key point to prevent biased models, by changing the optimization objective or imposing fairness constraints. In~\cite{Right_reason} the authors propose an adaptation of Domain Adaptation Neural Networks~\cite{DANN} to generate agnostic feature representations, unbiased related to certain protected attribute. Also based in domain adaption, in~\cite{wang2019racial} the authors reduce racial biases in face recognition using mutual information and unsupervised domain adaptation, from a labeled domain (i.e. Caucasian individuals) to an unlabeled one (i.e. non Caucasian individuals). A method to mitigate bias in occupation classification without having access to protected attributes is developed in~\cite{Bias_Bios_protected}, by reducing the correlation between the classifier's output for each individual and the word embeddings of their names. Wang \textit{et al.} studied in~\cite{wang2020mitigating} the use of an adaptive margin in large margin face recognition loss functions~\cite{ArcFace} to reduce the gap in performance between different ethnicity groups. They proposed to use deep Q-learning to adaptively find the margin for each demographic group during training. 

More recently, in-processing approaches based on adversarial learning frameworks~\cite{GAN} have been explored. A joint learning and unlearning method is proposed in \cite{Turning_Blind_Eye} to simultaneously learn the main classification task while unlearning biases by applying confusion loss, based on computing the cross entropy between the output of the best bias classifier and a uniform distribution. The authors of \cite{Learning_not_to_Learn} introduced a new regularization loss based on mutual information between feature embeddings and bias, training the networks using adversarial and gradient reversal \cite{DANN} techniques. In~\cite{SensitiveNets} an extension of triplet loss \cite{TripletLoss} is applied to remove sensitive information in feature embeddings, without losing performance in the main task. 

Finally, post-processing methods assume that the output of the model may be biased, so they apply a transformation on it to improve fairness between demographic groups. Some works in this line have proposed to prevent unfairness using discrimination-aware data-mining~\cite{Exploring_discrimination}\cite{DADM}. In~\cite{join_optimization}, the authors propose a framework that enables a human manager to select how to make the trade-off among fairness and utility. Then, the method selects a threshold for each demographic group to obtain an optimal classifier according to the manager's preferences. Post-processing techniques are also common among studies on fairness in ranking~\cite{yang2017measuring}\cite{celis2017ranking}\cite{zehlike2017fa}, which are close to our work here.

\subsection{Databases}
\label{sec:fairness_db}

The datasets used for learning or inference may be the most critical elements of the machine learning process where bias can appear. As these data are collected from society, they may reflect sociocultural biases~\cite{BigDataImpact}, or reflect an unbalanced representation of the different demographic groups composing it. A naive approach would be to remove all sensitive information from data, but this is almost infeasible in a general AI setup (e.g. \cite{Bias_in_Bios} demonstrates how removing explicit gender indicators from personal biographies is not enough to remove the gender bias from an occupation classifier, as other words may serve as ``proxy''). On the other hand, collecting large datasets that represent broad social diversity in a balanced manner can be extremely costly, and not enough to avoid disparate treatment between groups~\cite{wang2020mitigating}.

\begin{table*}
\resizebox{\textwidth}{!}{%
\centering
\def\arraystretch{1.5}
    \begin{tabular}{l|c|c|c|c|c|c}
        \hline
        \textbf{Database}&\textbf{\#Samples}&\textbf{Image}&\textbf{Text}&\textbf{Cat./Num.}&\textbf{Demographic}&\textbf{Access}\\
        \hline
        \hline
        UCI Adult Income ~\cite{UCI_rep}& $48.8$K&\xmark&\xmark&\cmark& Ethnicity, Gender& archive.ics.uci.edu/ml/datasets/adult \\
        \hline
        German Credit~\cite{UCI_rep}&$1$K &\xmark & \xmark &\cmark & Age & \makecell{archive.ics.uci.edu/ml/datasets/\\ statlog+(german+credit+data)}\\
        \hline
        Bank Marketing~\cite{2014bank}&$41.1$K & \xmark & \xmark & \cmark & Age& archive.ics.uci.edu/ml/datasets/Bank+Marketing\\
        \hline
        ProPublica Recidivism~\cite{Machine_bias}&$11.8$K & \xmark & \xmark & \cmark  & Ethnicity & github.com/propublica/compas-analysis\\
        \hline
        Common Crawl Bios~\cite{Bias_in_Bios}&$397$K & \xmark & \cmark & \xmark & Gender & github.com/microsoft/biosbias \\
        \hline
        WinoBias~\cite{WinoBias_dataset}&$3.2$K & \xmark & \cmark & \xmark & Gender & github.com/uclanlp/corefBias \\
        \hline
        CelebA~\cite{CelebA}&$202.6$K & \cmark & \xmark & \cmark & Gender & mmlab.ie.cuhk.edu.hk/projects/CelebA.html\\
        \hline
        IMDB-WIKI~\cite{IMDB_dataset}&$523$K & \cmark & \xmark & \xmark & Age, Gender & data.vision.ee.ethz.ch/cvl/rrothe/imdb-wiki/\\
        \hline
        Cleaned IMDB~\cite{Turning_Blind_Eye}& $140$K& \cmark & \xmark & \xmark & Age, Gender & robots.ox.ac.uk/~vgg/data/laofiw/ \\
        \hline
        MORPH~\cite{MORPH}& $55$K& \cmark & \xmark&\xmark &\makecell{Age, Ethnicity,\\ Gender} & \makecell{ebill.uncw.edu/C20231\_ustores/web/\\product\_detail.jsp?PRODUCTID=8} \\
        \hline
        PPB~\cite{Gender_shades}&$1.3$K &\cmark&\xmark&\xmark &Ethnicity, Gender & gendershades.org/index.html \\
        \hline
        LAOFIW~\cite{Turning_Blind_Eye}&$14$K&\cmark & \xmark & \xmark & Ethnicity, Gender & robots.ox.ac.uk/~vgg/data/laofiw/\\
        \hline
        FairFace~\cite{karkkainenfairface}& $108.5$K& \cmark& \xmark &\xmark & \makecell{Age, Ethnicity,\\ Gender} &github.com/joojs/fairface \\
        \hline
        Diversity in Faces~\cite{2019diversity}& $1$M & \cmark &\xmark &\xmark & \makecell{Age, Ethnicity, \\Gender }& \makecell{research.ibm.com/artificial-intelligence/\\trusted-ai/}\\
        \hline
        DiveFace~\cite{SensitiveNets}& $120$K & \cmark & \xmark & \xmark & Ethnicity, Gender & github.com/BiDAlab/DiveFace\\
        \hline
        BFW~\cite{2020bfw}& $20$K & \cmark & \xmark & \xmark & Ethnicity, Gender & github.com/visionjo/facerec-bias-bfw\\
        \hline
        DemogPairs~\cite{hupont2019demogpairs}&$10.8$K&\cmark&\xmark&\xmark&Gender, Ethnicity&\makecell{download.hertasecurity.com/research/\\DemogPairs.zip}\\
        \hline
        RFW~\cite{wang2019racial}& $40$K & \cmark &\xmark & \xmark& Ethnicity & whdeng.cn/RFW/testing.html\\
        \hline
        BUPT-B~\cite{wang2020mitigating}&$1.3$M & \cmark & \xmark & \xmark & Ethnicity & whdeng.cn/RFW/Trainingdataste.html\\
        \hline
        BUPT-G~\cite{wang2020mitigating}& $2$M& \cmark & \xmark & \xmark & Ethnicity & whdeng.cn/RFW/Trainingdataste.html \\
        \hline
        FairCVdb (Ours) & $24$K &\cmark&\cmark&\cmark& Ethnicity, Gender & github.com/BiDAlab/FairCVtest\\
        \hline

    \end{tabular}}
    \caption{Summary of the most common public databases for AI fairness and bias research. We specify the different modalities included in each dataset (i.e. images, texts, and categorical/numerical attributes), along with the demographic attributes typically studied with each one. }
    \label{tab:fairness_bbdd}
\end{table*}

The biases introduced in the dataset used to train machine learning models typically reflect human biases present in society, or are related to an inaccurate representation of groups~\cite{unbiased2011Torralba}\cite{serna2020sensitiveloss}. In view of this situation, the scientific community has put lots of effort into collecting databases that improve the representation of different demographic groups, which can be used to suppress the presence of bias. In this section, we discuss some of the most commonly used databases in AI fairness research, either because of the biases they present, or their absence (i.e. databases more balanced in terms of certain demographic attributes). Table~\ref{tab:fairness_bbdd} provides an overview of these databases, including the number of samples, data modality and the demographic attributes studied with each one.
The \textbf{Adult Income} dataset~\cite{UCI_rep} from the UCI repository is frequently used on gender and ethnicity bias mitigation. The main task of the database is predict whether a person will earn more or less than \$$50$K per year. The database includes $48,842$ samples with $14$ numerical/categorical attributes each, such as education level, capital-gain or occupation, and missing values.

The \textbf{German Credit} dataset~\cite{UCI_rep} contains $1$K entries with $20$ different categorical/numerical attributes, where each entry represents a loan applicant by a bank. The applicants are classified as good or bad risk credit, showing age bias toward young people. Also related to age biases, the \textbf{Bank Marketing} database~\cite{2014bank} contains marketing campaign data of a Portuguese bank institution. With more than $41$K samples, the goal is to predict if the client will subscribe a term deposit, based on $20$ categorical/numerical attributes including personal data and socioeconomic contextual information.

The \textbf{ProPublica Recidivism} dataset~\cite{Machine_bias} provides more than $11$K pretrial defendants records, assessed with the COMPAS algorithm to predict their likelihood of recidivism. After a $2$-year study, the researchers find out that the algorithm was biased towards African-Americans, showing both higher false positive and lower false negative rates than white defendants.

In the study of demographic bias in NLP technologies,\footnote{There are several works that study demographic biases in word embeddings~\cite{bolukbasi2016man}\cite{garg2018word}, working with representation spaces trained with large corpus of texts from Wikipedia, Common Crawl or Google News, among other sources.} we can cite the \textbf{Common Crawl Bios} dataset~\cite{Bias_in_Bios}, which contains nearly $400$K short biographies collected from Common Crawl. The goal of the dataset is to predict the occupation from these bios, out of $28$ possible occupations showing high gender imbalances. The dataset also provides a ``gender blinded'' version of each bio, where explicit gender indicators have been removed (e.g. pronouns or names). On a closely related task, the \textbf{WinoBias} database~\cite{WinoBias_dataset} provides $3,160$ sentences, where the goal is to find all the expressions related to certain entity. Centered in people entities referred by their occupations, the dataset requires to link gender pronouns to male/female stereotypical occupations.

We now focus in face datasets, which are the basis for different face analysis task such as face recognition or gender classification. The \textbf{CelebA} database~\cite{CelebA} contains nearly $202.6$K images from more than $10$K celebrities. Each image is annotated with $5$ facial landmarks, along with $40$ binary attributes including appearance features, demographic information, or attractiveness, which shows a strong gender bias.

The \textbf{IMDB-WIKI} dataset~\cite{IMDB_dataset} provides $460.7$K images from the IMDB profiles of $20,284$ different celebrities, along with $62.3$K images from Wikipedia. Images were labeled using the information available in the profiles (i.e. name, gender, and birth date), extracting an age label by comparing the timestamp of the images and the birth date. The dataset presents a gender bias in the age distributions, as we encounter younger females and older males. Due to the image acquisition process, some labels are noisy, so the authors of~\cite{Turning_Blind_Eye} released the \textbf{cleaned IMDB} dataset, with $60$K cleaned images for age prediction and $80$K for gender classification obtained from the IMDb split.

Also related with age studies, the \textbf{MORPH} database~\cite{MORPH} provides $55$K images from $13$K individuals, aimed to study the effect of age-progression on different facial tasks. The database is longitudinal with age, having pictures of the same user over time. The database is strongly unbalanced with respect to gender and ethnicity, with $65$\% images belonging to African-American males. 

Some databases aim to mitigate biases in face analysis technologies by putting emphasis in demographic balance and diversity. \textbf{Pilot Parliaments Benchmark (PPB)}~\cite{Gender_shades} is a dataset of $1,270$ parliamentarians images from $6$ different countries in Europe and Africa. The images are balanced with respect to gender and skin color, which are available as labels (the skin color is codified using the six-point Fitzpatrick system). The \textbf{Labeled Ancestral Origin Faces in the Wild (LAOFIW)} dataset~\cite{Turning_Blind_Eye} provides $14$K images manually divided into $4$ ancestral origin groups. The database is balanced with respect to ancestral origin and gender, and  a variety of pose and illumination. Also emphasizing ethnicity balance, the \textbf{FairFace} database~\cite{karkkainenfairface} contains more than $100$K images equally distributed in $7$ ethnicity groups (White, Black, Indian, East Asian, Southeast Asian, Middle East, and Latino), also providing gender and age labels. Aimed to study facial diversity, \textbf{Diversity in Faces}~\cite{2019diversity} provides $1$M images annotated with $10$ different facial systems including gender, age, skin color, pose, and facial contrast labels, among others.

If we look at face recognition databases, \textbf{DiveFace}~\cite{SensitiveNets} contains face images equitably distributed among $6$ demographic classes related to gender and $3$ ethnic groups (Black, Asian, and Caucasian), including $24$K different identities and a total of $120$K images. The \textbf{DemogPairs} database~\cite{hupont2019demogpairs} also proposes $6$ balanced demographic groups related to gender and ethnicity, each one with $100$ subjects and $1.8$K images. On his part, the \textbf{Balanced Faces in the Wild (BFW)} database~\cite{2020bfw} presents $8$ demographic groups related with gender and $4$ ethnicity groups (Asian, Black, Indian and White), each one with $100$ different users and $2.5$K images. Finally, Wang and Deng proposed three different databases based on MS-Celeb-$1$M~\cite{MS-Celeb}, namely \textbf{Racial Faces in the Wild (RFW)}~\cite{wang2019racial}, \textbf{BUPT-B}~\cite{wang2020mitigating} and \textbf{BUPT-G}~\cite{wang2020mitigating}. While RFW is designed as a validation dataset, aimed to measure ethnicity biases, both BUPT-B and BUPT-G are proposed as ethnicity-aware training datasets. RFW defines $4$ ethnic groups (Caucassian, Asian, Indian, and African), each one with $10$K images and $3$ different subjects. On the other hand, both BUPT-B and BUPT-G propose the same ethnic groups, the first one almost ethnicity-balanced with $1.3$M images and $28$K subjects, while the latter contains $2$M images and $38$K subjects, which are distributed approximating the world's population distribution.


\section{FairCVdb: Dataset for Multimodal Bias Research}
\label{sec:FairCVdb}

\subsection{AI in Hiring Processes}\label{sec:Framework}

The usage of predictive tools in recruitment processes is increasing nowadays. Employers have adopted these tools in an attempt to reduce the time and cost of hiring, or to maximize the quality of the hiring process, among other reasons~\cite{Hiring_Algorithms_Report}. Rather than a single-point decision, the hiring pipeline suppose a multi-stage process, which can be broadly divided in four stages~\cite{Hiring_Algorithms_Report}. In the \textit{sourcing} stage the employers attract potential candidates through advertisements or job posting. Then, during \textit{screening} the employers assess the applicants to choose a subset to \textit{interview} individually. Finally, employers make a final decision (i.e. whether to hire or reject each applicant) in the \textit{selection} stage. All of these stages can benefit from the use of automatic algorithms\footnote{https://www.hirevue.com/}, as well as suffer from algorithmic discrimination if systems are not carefully designed. The labor market has a long history of unfair treatment of minority groups~\cite{bertrand2004emily}\cite{bendick1997employment}, which makes bias prevention a crucial step in automatic hiring tools design.
Although the study of fairness in algorithmic hiring has been limited~\cite{schumann2020we}, some works are starting to address this topic~\cite{raghavan2020mitigating}\cite{sanchez2020does}\cite{cowgill2018bias}. 

For the purpose of studying discrimination in Artificial Intelligence at large, and particularly in hiring processes, in this work we propose a new experimental framework inspired in a fictitious automated recruiting system: FairCVtest. Our work can be framed within the screening stage of the hiring pipeline, where an automatic tool determines a score from a set of applicants resumes. We chose this application because it comprises personal information from different nature \cite{2018_INFFUS_MCSreview1_Fierrez}. 

\begin{figure*}[t!]
\centering
\includegraphics[width = 0.85\textwidth]{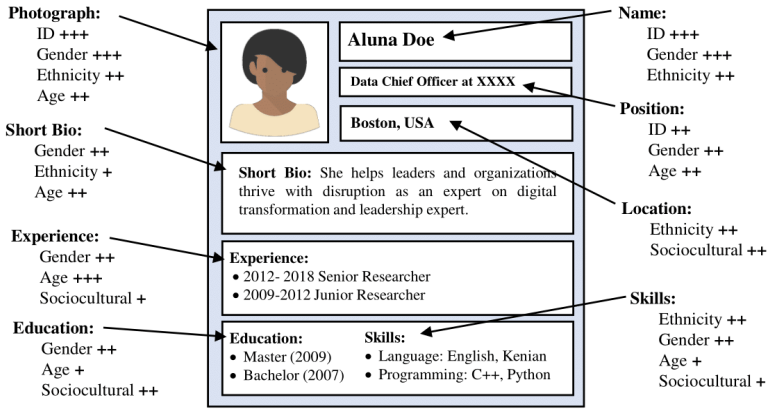}
\caption{Information blocks in a resume and personal attributes that can be derived from each one. The number of crosses represent the level of sensitive information (+++ =  high, ++ = medium, + = low).}
\label{fig:cv}
\end{figure*}

The resume is traditionally composed by structured data including name, position, age, gender, experience, or education, among others (see Figure \ref{fig:cv}), and also includes unstructured data such as a face photo or a short biography. A face image is rich in unstructured information such as identity, gender, ethnicity, or age \cite{2018_TIFS_SoftWildAnno_Sosa}\cite{Understanding_faces}. That information can be recognized in the image, but it requires a cognitive or automatic process trained previously for that task. The text is also rich in unstructured information. The language and the way we use that language, determine attributes related to your nationality, age, or gender. Both, image and text, represent two of the domains that have attracted major interest from the AI research community during last years. The Computer Vision and the Natural Language Processing communities have boosted the algorithmic capabilities in image and text analysis through the usage of massive amounts of data, large computational capabilities (GPUs), and deep learning techniques. 

The resumes used in the proposed FairCVtest framework include merits of the candidate (e.g. experience, education level, languages,  etc.), two demographic attributes (gender and ethnicity), and a face photograph (see Section \ref{sec:Dataset} for all the details). 

\subsection{FairCVdb: Dataset Description}
\label{sec:Dataset}

In this work we present FairCVdb, a new dataset with $24,000$ synthetic resume profiles for both fairness and multimodal research in AI. Each profile includes $2$ demographic attributes (gender and ethnicity), an occupation, a face image, a name, $7$ attributes obtained from $5$ information blocks that are usually found in a standard resume, and a short biography. The profiles comprise data from different nature including structured and unstructured data: 

\begin{itemize}
    \item Demographic attributes (\textit{structured data)}: Each profile has been generated according to two gender classes and three ethnicity classes. These demographic attributes determine the face image (gender and ethnicity related), name (gender related), and pronouns in the short biography (gender related). 
    \item Face image (\textit{unstructured data - image}): Each profile contains a real and unique face image assigned from the DiveFace database~\cite{SensitiveNets}, which was introduced in Section~\ref{sec:fairness_db}. DiveFace\footnote{https://github.com/BiDAlab/DiveFace} contains  face images from $24$K different identities with their corresponding gender and ethnicity attributes. 
    \item Short Biography (\textit{unstructured data - text}):  We use the Common Crawl Bios dataset~\cite{Bias_in_Bios} to associate a short biography, a name, and an occupation (from a pool of $10$ different occupations) to each profile. 
    \item Candidate competencies (\textit{structured data}): The $5$ information blocks are: $1$) education attainment, $2$) availability, $3$) previous experience, $4$) the existence of a recommendation letter, and $5$) language proficiency in a set of $3$ different and common languages. Each language is encoded with an individual feature ($3$ features in total) that represents the level of knowledge in that language. We will refer to these resume features as candidate competencies. 

\end{itemize}

As we previously mentioned in Section~\ref{sec:fairness_db}, the Common Crawl Bios dataset\footnote{https://github.com/microsoft/biosbias} ~\cite{Bias_in_Bios}   contains online biographies collected from Common Crawl relating $28$ different occupations. Gender and occupation labels are available for each biography, as well as a ``blinded'' version of the bio, in which explicit gender indicators have been removed. For example, a biography labeled as [Attorney, Female] is presented as: \textit{\underline{Andrea Jepsen} is an attorney with the School Law Center, a law firm focusing on the rights of students and families in education and school law disputes. \underline{She} has worked with people with disabilities since 1997 in a variety of roles, including as an early childhood special education service coordinator, and as a legal services provider working regularly in the courts and in administrative proceedings. \underline{Ms. Jepsen}’s broad legal experience has involved representing clients in a variety of critical legal issues related to education, housing, elder law matters, public benefits, family law disputes, probate and other concerns.} Note that we underlined explicit gender indicators removed in the ``blinded bio'', and that both name and occupation can be found in the first sentence of each biography, so this sentence was not included in the bios.

We select $24$K biographies, and its corresponding blinded versions, from a subset of $10$ different occupations. Each biography is associated according to gender to one FairCV profile, providing as well an occupation label and a name to the profiles, which we obtain by processing the first sentence of each bio. We group the occupations in $4$ professional sectors: $1$) audiovisual communication and journalism, with \textit{journalist, photographer}, and \textit{filmmaker}; $2$) administration and jurisdiction, with \textit{attorney} and \textit{accountant}; $3$) healthcare, with \textit{surgeon, nurse}, and \textit{physician}; and $4$) education, with \textit{professor} and \textit{teacher}. Each professional sector has the same number of samples (i.e. $6$K bios), and is gender-balanced. Furthermore, we define a suitability attribute ($S$), representing the affinity degree of each sector with the potential job to which the resumes apply. The association of this attribute with each sector has purely academic purposes, without seeking to state the usefulness or importance of each of them.

The score $T^j$ for a profile $j$ is generated by linear combination of the candidate competencies $\textbf{x}^j = [x^j_1, ..., x^j_n]$ and the suitability attribute $S^j$ as:

\begin{equation}
\label{eqn:Score_gen}
    T^j = \beta^j + \sum_{i = 1}^{n} \alpha_{i} x^j_i + \alpha_{s} S^j  
\end{equation}

\noindent where $n = 7$ is the number of features (competencies), $\alpha_i$ are the weighting factors for each competency $x_i^j$ (fixed manually based on consultation with a human recruitment expert), and $\beta^j$ is a small Gaussian noise to introduce a small degree of variability (i.e. two profiles with the same competencies do not necessarily have to obtain the same result in all cases). Those scores $T^j$ will serve as groundtruth in our experiments.

Note that, by not taking into account gender or ethnicity information during the score generation in Equation~(\ref{eqn:Score_gen}), these scores become agnostic to this information, and should be equally distributed among different demographic groups. Thus, we will refer to this target function as Unbiased scores $T^U$, from which we define two target functions that include two types of bias: Gender bias $T^G$ and Ethnicity bias $T^E$. Biased scores are generated by applying a penalty factor $T_\delta$ to certain individuals belonging to a particular demographic group. For the Gender-biased scores $T^G$, we apply a penalty factor on the female group, while in the Ethnicity-biased scores $T^E$ we apply the penalty factor to one ethnic group, and the inverse to another one (i.e. the individuals belonging to this group are overrated in $T^E$, showing a higher score than in $T^U$). This leads to a set of scores where, with the same competencies, certain groups have lower scores than others, simulating the case where the process is influenced by certain cognitive biases introduced by humans, protocols, or automatic systems. 

\begin{table}[t]
    \centering
    \begin{tabular}{c|c|c}
    \hline
         \textbf{Name}& \textbf{Type}& \textbf{Data values}\\
         \hline\hline
         Education& I&$x_1\in\{0.2,0.4,0.6,0.8,1\}$ \\
         \hline
         Recommendation&I& $x_2\in\{0,1\}$\\
         \hline
         Availability&I&$x_3\in\{0.2,0.4,0.6,0.8,1\}$\\
         \hline
         Previous experience&I&$x_4\in\{0,0.2,0.4,0.6,0.8,1\}$\\
         \hline
         Language proficiency&I&\makecell{$x_i\in\{0,0.2,0.4,0.6,0.8,1\},$\\$i\in\{5,6,7\}$}\\
         \hline
         Face Image&I&I$\,[m,n]\slash\, m,n\in[0,119]$\\
         \hline
         Face embedding&I&$\boldsymbol{\mathrm{f}}\in\mathbb{R}^{20}\slash\|\boldsymbol{\mathrm{f}}\| = 1$\\
         \hline
         Agnostic face embedding&I&$\boldsymbol{\mathrm{f}}_a\in\mathbb{R}^{20}\slash\|\boldsymbol{\mathrm{f}}_a\| = 1$\\
         \hline
         Name&I& Text data\\
         \hline
         Biography&I& Text data\\
         \hline
         Agnostic Biography&I&Text data\\
         \hline
         Gender&I/T&$G\in\{0,1\}$\\
         \hline
         Ethnicity&I/T&$E\in\{0,1,2\}$\\
         \hline
         Occupation&I/T&$O:\mathbb{N}\in[0,11]$\\
         \hline
         Suitability&I/T&$S\in\{0.25,0.5,0.75,1\}$\\
         \hline
         Blind score&T&$T^U:\mathbb{R}\in[0,1]$\\
         \hline
         Gender biased score&T&$T^G:\mathbb{R}\in[0,1]$\\
         \hline
         Ethnicity biased score&T&$T^E:\mathbb{R}\in[0,1]$\\
         \hline

    \end{tabular}
    \caption{Overview of the different attributes available in each FairCV profile. We include the possible values of each attribute, as well as its nature as Input and/or Target.}
    \label{tab:faircvdb}
\end{table}

\begin{figure}[t!]
\centering
\includegraphics[width=0.85\columnwidth]{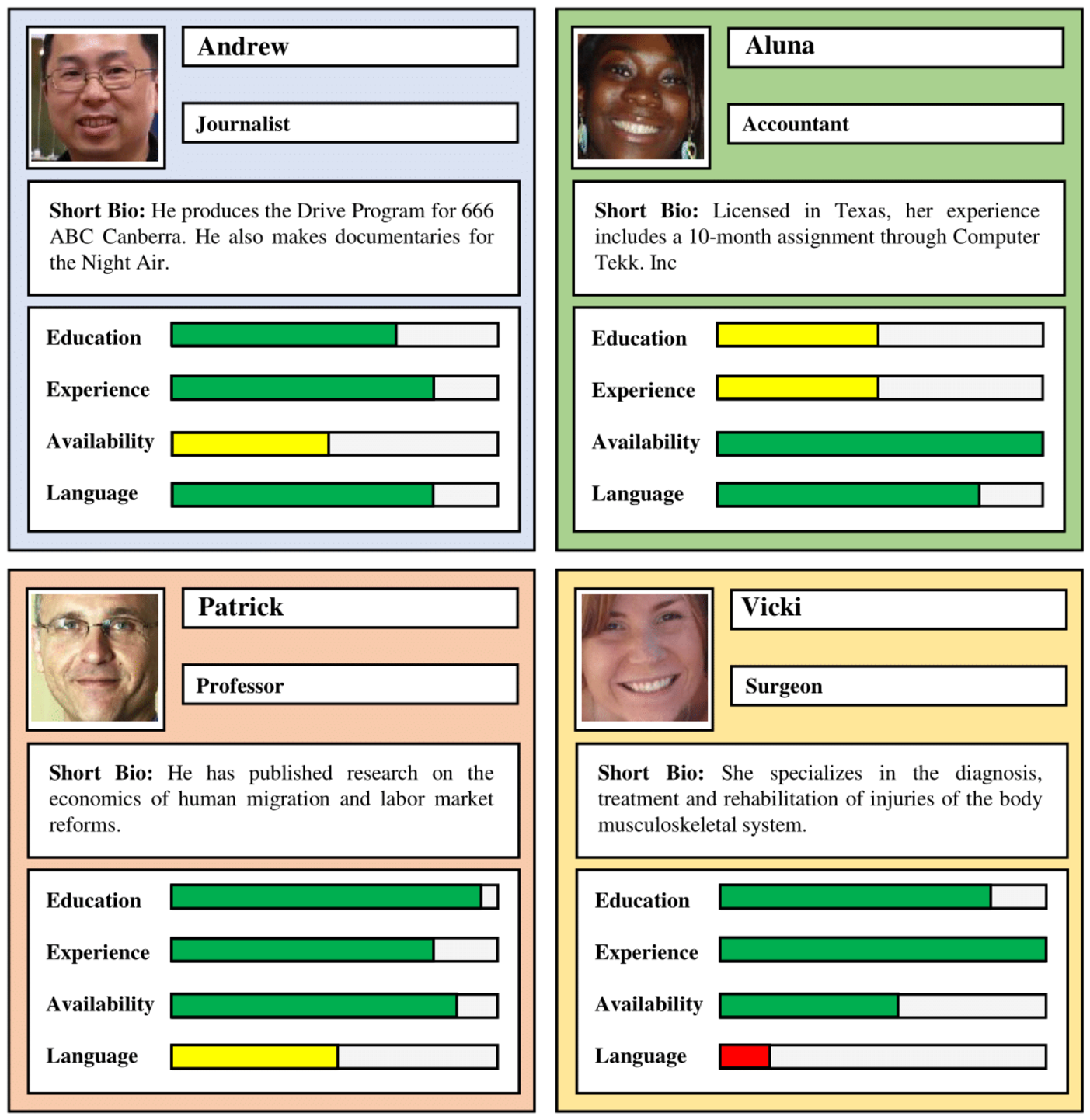} 
\caption{Visual examples of the FairCVdb synthetic resumes, including a face image, a name, an occupation, a short biography and the candidate competencies.}
\label{fig:CV_examples}
\end{figure}

Table~\ref{tab:faircvdb} summarizes the features that make up each profile, as well as their labels. We divided the FairCVdb in two splits, with $80$\% of the synthetic profiles ($19,200$ CVs) as training set, and the remaining $20$\% ($4,800$ CVs) as validation set. Both sets are almost perfectly balanced among gender, ethnicity and professional sector. Fig.~\ref{fig:CV_examples} presents four visual examples of the resumes generated with FairCVdb.

\begin{figure*}[t]
\centering
\includegraphics[width=0.85\textwidth]{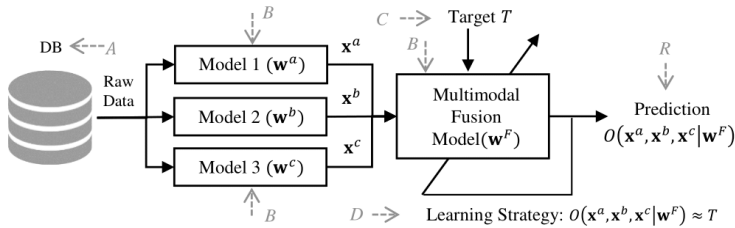} 
\caption{Block diagram of the automatic multimodal learning process and $6$ (\emph{A} to \emph{E}) stages where bias can appear.}
\label{fig:Block_diagram}
\end{figure*}

\section{FairCVtest: Description }\label{sec:description}

\subsection{General Learning Framework}
\label{sec:problem_formulation}

The multimodal model represented by its parameters vector $\boldsymbol{\mathrm{w}}^{F}$ ($F$ for fused model~\cite{2018_INFFUS_MCSreview1_Fierrez}) is trained using features learned by $M$ independent models $\{\boldsymbol{\mathrm{w}}^{1},...,\boldsymbol{\mathrm{w}}^{M}\}$ where each model produces $n_i$ features $\boldsymbol{\mathrm{x}}^{i} = [x^{i}_{1},...,x^{i}_{n_i}] \in \mathbb{R}^{n_i}$. Without loss of generality, the Figure \ref{fig:Block_diagram}  presents the learning framework for $M=3$. 
 The learning process is guided by a Target function $T$, and a learning strategy that minimizes the error between the output $O$ and the Target function $T$. In our framework, where $\boldsymbol{\mathrm{x}}^{i}$ is data obtained from the resume, $\boldsymbol{\mathrm{w}}^{i}$ are models trained specifically for different information domains (e.g. images, text) and $T$ is a score within the interval [$0$, $1$] ranking the candidates according to their merits. A score close to $0$ corresponds to the worst candidate, while the best candidate would get $1$.  The learning strategy is traditionally based on the minimization of a loss function defined to obtain the best performance. The most popular approach for supervised learning is to train the model $\boldsymbol{\mathrm{w}}$ by minimizing a loss function $\mathcal{L}$ over a set of training samples $\mathcal{S}$:

\begin{equation}
\label{eqn:learning_strategy}
    \min_{\boldsymbol{\mathrm{w}}^{F}}{\sum_{\boldsymbol{\mathrm{x}}^{j} \in \mathcal{S}}\mathcal{L}(O(\boldsymbol{\mathrm{x}}^j\mid\boldsymbol{\mathrm{w}}^{F}),T^j)} 
\end{equation}

Biases can be introduced in different stages of the learning process (see Figure \ref{fig:Block_diagram}): in the Data used to train the models (\textit{A}), the Preprocessing or Feature generation (\textit{B}), the Target function (\textit{C}), and the Learning strategy (\textit{D}). As a result of the biases introduced at of these points ($A$ to $D$), we may obtain biased Results (\textit{R}). In this work we focus on the Target function (\textit{C}) and the Learning strategy (\textit{D}). The Target function is critical as it could introduce cognitive biases from biased processes.


\subsection{FairCVtest: Multimodal Learning Architecture for Automatic CV Analysis}

\begin{figure*}[t]
\centering
\includegraphics[width=1\textwidth]{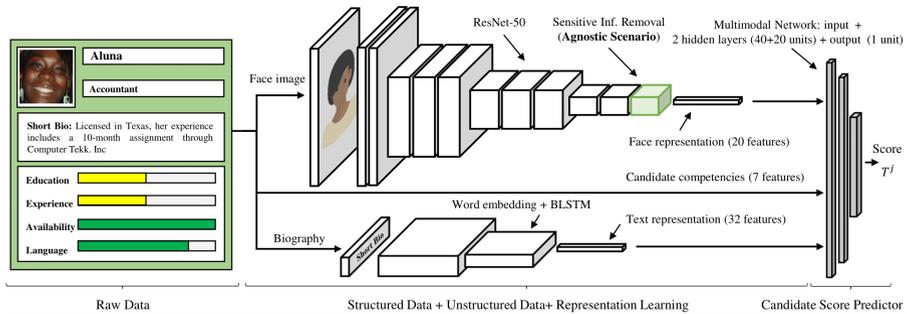} 
\caption{Multimodal learning architecture, composed by a Convolutional Neural Network (ResNet-$50$~\cite{he2015deep}), a BLSTM and a fully connected network used to fuse the features from different domains (image, text and structured data). Note that, in the agnostic scenario, we include a sensitive information removal module~\cite{SensitiveNets} over the ResNet network to generate agnostic face embeddings.}
\label{fig:network}
\end{figure*}

Figure \ref{fig:network} summarizes the learning architecture proposed to study the different scenarios of FairCVtest. We designed the candidate score predictor as a multimodal neural network with three input branches: i) face image, ii) text biography, and iii) candidate competencies. The learning architecture includes two specific models to process the face image and text data from the biography, before fusing the information from all three modalities.

\subsubsection{Face Analysis Model}

We use the face image from each profile, and the pre-trained model ResNet-$50$~\cite{he2015deep} as feature extractor to obtain feature embeddings from the applicants' face attributes. ResNet-$50$ is a popular Convolutional Neural Network composed with $50$ layers including residual or ``shortcuts'' connections to improve accuracy as the net depth increases (i.e. solving the ``vanishing gradient'' problem). ResNet-$50$'s last convolutional layer outputs embeddings with $2048$ features, so we added a fully connected layer to perform a bottleneck that compresses these embeddings to just $20$ features (maintaining the competitive face recognition performance), so that its size approximates to that of the candidates competencies. Note that our face model was trained exclusively for the task of face recognition. However, although gender or ethnicity information were not intentionally employed during the training process, this information is part of the face attributes. Therefore, an AI system trained on these face embeddings could detect the protected attributes without being explicitly trained for this task. 

\subsubsection{Text Analysis Model}

The second branch is aimed to extract a text representation from the bios, using a bidirectional LSTM layer composed by $32$ units and hyperbolic tangent activation. This branch receives as input a sequence of word vectors. We use the fastText\footnote{https://fasttext.cc/docs/en/english-vectors.html} word embeddings~\cite{mikolov2018advances} to represent each word in the biographies as $300$-dimensional word vectors. Note that these word vectors were trained on a different Common Crawl subset than the one used to extract the biographies of~\cite{Bias_in_Bios}. 

\subsubsection{Multimodal Model}

The face and text features obtained from its respective models are combined with the candidate competencies to feed the multimodal network. This network is composed by two hidden layers, with $40$ and $20$ neurons respectively and ReLU activation, and only one neuron with sigmoid activation in the output layer. Note that, as the target functions $T$ in FairCVdb are real valued scores within the interval [$0$, $1$], we treat this task as a regression problem. A binary classifier can be obtained by thresholding the predicted scores (i.e. switching from a scoring tool to a selection tool), as we will show in Section~\ref{sec:selection}.

\subsubsection{Privacy-enhancing Representation Learning}

With the aim of generating another representation, agnostic with regard to gender and ethnicity, we use the method proposed in~\cite{SensitiveNets}, called SensitiveNets. This method was proposed to improve the privacy in face biometrics, by incorporating an adversarial regularizer capable of removing sensitive information from pre-trained feature embeddings without losing performance in the main task. Thus, two different face representations are available for each profile, one containing gender and ethnicity sensitive information, and a second one ``blind'' or agnostic to these attributes. In order to remove sensitive information from the learned space, the equation \ref{eqn:learning_strategy} is replaced by:

\begin{equation}
\label{eqn:sensitivenets}
    \min_{\boldsymbol{\mathrm{w}}^{F}}{\sum_{\boldsymbol{\mathrm{x}}^{j} \in \mathcal{S}}\mathcal{L}(O(\boldsymbol{\mathrm{x}}^j\mid\boldsymbol{\mathrm{w}}^{F}),T^j)+\Delta} 
\end{equation}

\noindent where $\Delta$ is an adversarial regularizer introduced to measure the amount of sensitive information available in the learned space represented by $\boldsymbol{\mathrm{w}}^{j}$:

\begin{equation}
\label{delta}
     \Delta=\log\{ \, 1 + \mid 0.9 - P(\,Male \,\mid\,\boldsymbol{\mathrm{x}}^{j})\mid\} 
\end{equation}

The probability $P$ is the output of a classifier trained to detect the sensitive attribute in the learned space (e.g., Gender in this example). In other words, $P$ is the probability of observing  \emph{Male} features in the learned space after the sensitive information suppression (see~\cite{SensitiveNets} for details).

\subsection{Scenarios and Protocols}\label{sec:protocol}
In order to evaluate how and to what extent an algorithm is influenced by biases that are present in the FairCVdb target function, we use the FairCVdb dataset previously introduced in Section \ref{sec:FairCVdb} to train a recruitment system under $3$ different scenarios. The proposed testbed (FairCVtest) consist of FairCVdb, the trained recruitment systems, and the related experimental protocols.

We present $3$ different versions of the recruitment system, with slight differences in the input data and the target function aimed at studying gender/ethnicity biases in multimodal learning. The $3$ scenarios included in FairCVtest were all trained using the candidate competencies, a face representation, and a short bio, with the following particular configurations:

\begin{itemize}
\setlength\itemsep{0em}
    \item \textit{Neutral:} Training with Unbiased scores $T^U$, the original face representation extracted with ResNet-$50$~\cite{he2015deep}, and the biography with explicit gender indicators.
    \item \textit{Biased:} Training with Biased scores $T^{(G/E)}$, the original face representation, and the biography with explicit gender indicators.
    \item \textit{Agnostic:} Training with Biased scores $T^{(G/E)}$, the gender and ethnicity agnostic representation learned with~\cite{SensitiveNets}, and the ``blind'' biography.
    
\end{itemize}

 The experiments performed in next section will try to evaluate the capacity of the recruitment AI in each scenario to detect protected attributes (e.g. gender, ethnicity) without being explicitly trained for this task.

\section{Experiments and Results}\label{sec:experiments}

In this section we will train and evaluate different recruitment models, aimed to predict a score from the candidate resumes. Each recruitment tool follows the configuration of one of the scenarios exposed in Section~\ref{sec:protocol}, and was trained for $16$ epochs using Adam optimizer ($\alpha = 0.001$, $\beta_1= 0.9$ and $\beta_2 = 0.999$), batch size of $128$, and mean absolute error as loss metric.
\begin{figure*}[t]
\centering
\includegraphics[width = \textwidth]{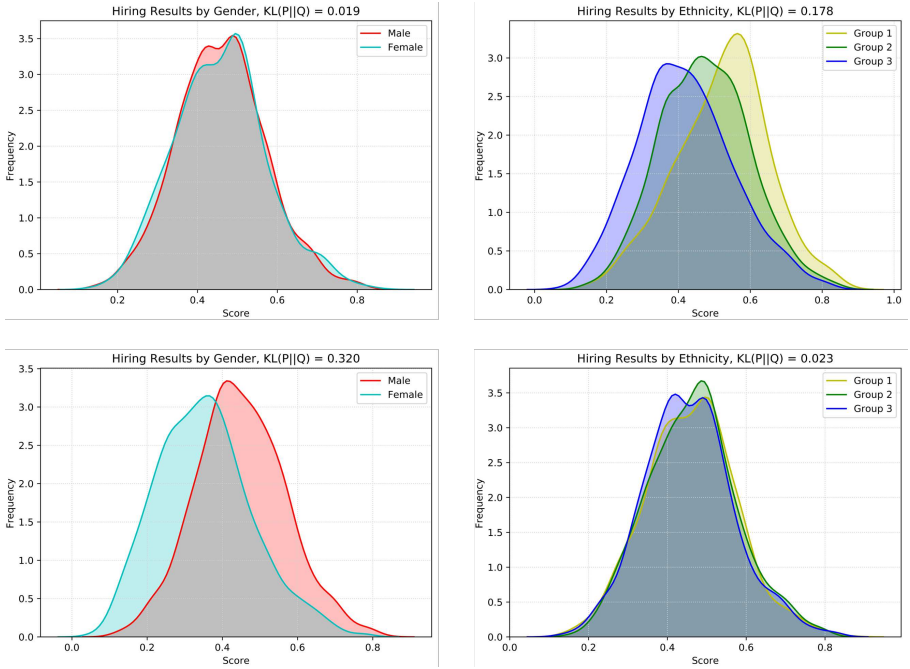}
\label{fig:demo_distributions}
\caption{Hiring score distributions by gender (left) and ethnicity (right). The top row presents hiring score distributions in the Neutral Scenario, while the bottom presents them in the Gender- and Ethnicity-biased Scenarios.}
\end{figure*}

In Figure~\ref{fig:demo_distributions} we can observe the distribution of the scores predicted from our validation set, by gender or ethnicity, in both Neutral and Biased scenarios. As a measure of the bias' impact in the classifier, we compute the Kullback-Leiber divergence KL($P\,\|\,Q$) between demographic distributions. In the gender case, we define $P$ as the male score distribution and $Q$ as the female's one, while in the ethnicity setup we make $1$-$1$ comparisons (i.e. G$1$ vs G$2$, G$1$ vs G$3$ and G$2$ vs G$3$) and report the average divergence. In the Neutral Scenario (see top row in Figure~\ref{fig:demo_distributions}) there is no difference between demographic groups, as can be corroborated with the KL divergence tending to zero in both cases (KL =$0.019$ in the gender case, KL = $0.023$ in the ethnicity one). As expected, using the unbiased scores $T^U$ as target function and a balanced training set leads us to an unbiased classifier, even in the presence of data containing demographic information (as we will see in Section~\ref{sec:sensitive_information}). On the other hand, the demographic difference is clearly visible in the Biased scenarios. This difference is most notorious in the gender case (see bottom-left plot in Figure~\ref{fig:demo_distributions}), with the KL divergence rising to $0.320$, compared to its low value in the Neutral setup. Attending to the Ethnicity-biased Scenario, the average KL divergence rises to $0.178$. However, the difference between Groups $1$ and $3$ is close to that seen between male-female classes, with a KL divergence around $0.317$. Note that gender or ethnicity are not inputs of our model, but rather the system is able to detect this sensitive information from some of the input features (i.e. the face embedding, the biography, or the competencies). Therefore, despite not having explicit access to demographic attributes, the classifier is able to detect this information and find its correlation with the biases introduced in the scores, and so it ends up reproducing them.

\begin{figure}[t]
    \centering
    \includegraphics[width = \textwidth]{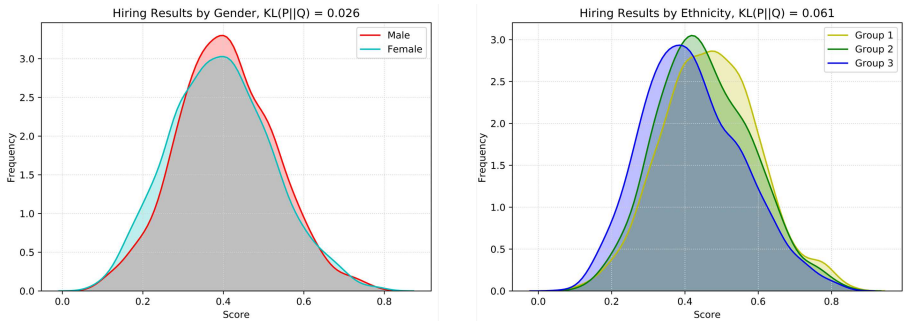}
    \label{fig:agnostics_distributions}
    \caption{Hiring score distributions by gender (left) and ethnicity (right), in the Agnostic Scenario.}
\end{figure}

The third scenario provided by FairCVtest, which we call Agnostic Scenario, aims to prevent the system to inherit data biases. As we introduced in Section~\ref{sec:protocol}, the Agnostic Scenario uses a gender blind version of the biographies, as well as a face embedding where sensitive information has been removed using the method of ~\cite{SensitiveNets}. Figure~\ref{fig:agnostics_distributions} presents the hiring score distributions in this setup. As we can see, the gender distributions are close to the ones observed in the Neutral Scenario (see top-left plot in Figure~\ref{fig:demo_distributions}), despite using gender-biased labels during training. In the ethnicity case, we can observe a slight difference between groups, much smoother than the one we saw in the Biased Scenario (see bottom-left plot in Figure~\ref{fig:demo_distributions}), as can be confirmed with the KL divergence (i.e. $0.061$, compared
to the biased case where this value is around $0.178$). However, this gap on the scores between demographic groups still has margin to decrease to a level similar to that of the Neutral Scenario. The difference observed in the behavior of gender and ethnicity agnostic cases can be explained by the fact that we removed almost all gender information from the input (i.e. face embedding and biography), but for the ethnicity we only took measures on the face embedding, not on the competencies. Thus, competencies are acting as a soft proxy for the ethnicity group. 

Note that our agnostic approximation does not seek to make the system capable of detecting whether a score is unfair, nor to compensate such bias, but rather blind it to sensitive attributes with the aim of preventing the model to establish a correlation between the demographic groups and score biases. This fact can be corroborated with the training loss, which has a higher value in the Agnostic Scenario ($0.035$ for gender, $0.044$ for ethnicity) than in the Biased Scenario ($0.49$ for gender, $0.64$ for ethnicity). By removing sensitive information from the input, the model is not able to learn what motivates the difference in the scores between individuals with similar competencies, as it is blind to the demographic group, and therefore its output does not approximate correctly the biased target function after training.

\subsection{Fairness in Recruitment Tools: Learning Demographic Parity}
\label{sec:selection}

Now that we have analyzed the effect of data biases in the score distributions, in this section we evaluate their impact in the final decision of a screening process. A screening tool is used to assess a set of individuals according to certain criteria to select a subset of the ``best'' ones. The outcome of such process could be a list of selected candidates (e.g. applicants selected for an individual interview) or a top-$k$ ranking that measures the relative quality of the $k$ best individuals from the set. We propose an experiment to simulate a screening process with FairCtest, using the recruitment tools that we trained in the previous section. For each scenario, we predict the scores from a pool including the $4,800$ resumes of our validation set, and select the top-$1000$ candidates (i.e. the candidates with the highest scores) among them. By selecting the $1000$ candidates with the highest scores, we establish a thresholding rule to classify the candidates in two categories, therefore switching from a regression task to a binary classification task.

We will measure fairness in each scenario using the demographic parity criterion. This criterion requires a classifier's decision  to be statistically independent of a protected attribute (i.e. gender or ethnicity in our experiments). As we're working with balanced groups, the criterion implies that all  demographic groups should have the same rate of appearance in the top. We can measure demographic parity between two groups through the $p\%$ score as:

\begin{equation}
p\% = \min \left(\frac{P(\hat{y} = 1\mid z = 0)}{P(\hat{y} = 1\mid z = 1)},\frac{P(\hat{y} = 1\mid z = 1)}{P(\hat{y} = 1\mid z = 0)}\right)
    \label{eqn:p_score}
\end{equation}

\noindent where $\hat{y}$ is a trained classifier's prediction, and $z$ is a binary protected attribute. The p$\%$ score calculates how far off the equality the model's decisions are. According to the U.S. Equal Employment  Opportunity Commission ``$\slfrac{4}{5}$ rule''~\cite{biddle2006adverse}, the positive rate of a protected group shouldn't be less than $\slfrac{4}{5}$ of that of the group with the higher positive rate. Otherwise, the protected group could be suffering disparate impact. Hence, we will set this rate as an indicator that a model is biased.

Table~\ref{tab:top_scores} presents the top-$1000$ candidates in each scenario, by gender and ethnicity group. In the ethnicity case, we compute three $p\%$ scores per model by doing $1$-$1$ comparisons with the three ethnic groups. As we can observe, in the Neutral Scenario the classifier shows no demographic bias, with both gender and ethnicity groups having a balanced representation in the ranking. This can be corroborated with the p$\%$ score, which reach values higher than $90\%$ in all cases. In the Biased scenario the groups Male and Ethnic Group $1$ are significantly favored and the difference between groups is now clearly visible. In the gender case, almost $70\%$ of the individuals in the top belong to the Male group, which reduces the $p\%$ score nearly to $40\%$. On the other hand, the first ethnic group represents almost half of the top, with the third one exhibiting a $21.6\%$. For both $G2$ and $G3$ the $p\%$ score points out unfair treatment (i.e. a value under $80\%$) with respect to G$1$ (see $p1\%$ and $p2\%$ in Table~\ref{tab:top_scores}). Finally, in the Agnostic Scenario the demographic differences were significantly reduced with respect to the Biased one, with Male and Female rates showing even more balance than in the Neutral Scenario. The reduction of the gap among ethnic rates is enough to overcome the limit in the $p\%$ score, but still leaves room for improvement with a difference of nearly $6\%$ between G$1$ and G$3$. This is not surprising, as we already observed in Figure~\ref{fig:agnostics_distributions} an slight difference between the score distribution of each ethnicty group.

\begin{table*}[t]
\resizebox{\textwidth}{!}{%
    \centering
    \begin{tabular}{l|c|c|c|c|c|c|c|c|c}
    \hline
    \multirow{2}{*}{\textbf{Scenario}}&\multicolumn{2}{c|}{\textbf{Gender}}&\multirow{2}{*}{$\boldsymbol{p\%}$}&\multicolumn{3}{c|}{\textbf{Ethnicity}}&\multirow{2}{*}{$\boldsymbol{p_1\%}$}&\multirow{2}{*}{$\boldsymbol{p_2\%}$}&\multirow{2}{*}{$\boldsymbol{p_3\%}$}\\
    \cline{2-3}\cline{5-7}
    &\textbf{Male}&\textbf{Female}&&\textbf{Group 1}&\textbf{Group 2}&\textbf{Group 3}&&&\\
    \hline
    \hline
    Neutral&$51.90\%$&$48.10\%$&$92.68\%$&$34.20\%$&$35.00\%$&$30.80\%$&$97.71\%$&$90.06\%$&$88.00\%$\\
    \hline
    Biased&$72.90\%$&$27.10\%$&$37.17\%$&$50.80\%$&$30.40\%$&$18.80\%$&$59.84\%$&$37.01\%$&$61.84\%$\\
    \hline
    Agnostic&$52.80\%$&$47.20\%$&$89.39\%$&$36.70\%$&$32.70\%$&$30.60\%$&$89.10\%$&$83.38\%$&$93.58\%$\\
    \hline
    \end{tabular}}
    \caption{Distribution of the top 1000 candidates in each Scenario of FairCVtest, by gender and ethnicity group. We include the $p\%$ score (see Eqn.~\ref{eqn:p_score}) as a measure of the difference between groups. In the ethnicity case, $p_1\%$ for G$1$ vs G$2$, $p_2\%$ for G$1$ vs G$3$ and $p_3\%$ for G$2$ vs G$3$.}
    \label{tab:top_scores}
\end{table*}

\subsection{Privacy in recruitment tools: Removing sensitive information}\label{sec:sensitive_information}

We have observed in the previous sections the impact of demographic biases in both the score distribution and the selection rates in different scenarios. In these experiments, the difference between groups was a consequence of the biases introduced in the target function. However, as can be seen in the Agnostic Scenario, by removing gender and ethnicity information from the input we can prevent the model to reproduce those biases, as it cannot \emph{see} which factor determines the score penalty for some individuals. 

Since the key of our Agnostic Scenario is the removal of sensitive information, in this section we will analyze the demographic information extracted by the hiring tool in each scenario. To this aim, we use multimodal feature embeddings extracted by the recruitment tool to train and evaluate the performance of both gender and ethnicity classifiers. We obtain these embeddings as the output of the first dense layer of our learning architecture (see Section~\ref{sec:protocol}), in which the information from different data domains has already been fused. For each scenario, we train $3$ different classification algorithms, namely Support Vector Machines (SVM), Random Forests (RF), and Neural Networks (NN).

Table~\ref{tab:demographics} presents the accuracies obtained by each classification algorithm in the $3$ scenarios of FairCVtest. The results show a different behavior between scenarios and demographic traits. As expected, the setup in which most sensitive information can be extracted (gender and ethnicity in this work) is the Biased one for both attributes. The SVM classifier obtains the higher validation accuracies, with almost $90\%$ in the gender case and $76.40\%$ in the ethnicity one. Note that none of these values reach state-of-art performances (i.e. neither the ResNet-$50$ model nor the hiring tools weren't explicitly trained to classify those attributes), but both of them warn of large amounts of sensitive information within the embeddings. On the other hand, both Neutral and Agnostic scenarios show lower accuracies than the Biased configuration. However, we can see a gap in performance between them, with all the classifiers showing higher accuracy in the Neutral Scenario. This fact demonstrates that, despite training with the Unbiased scores $T^U$ which have no relationship with any demographic group membership, the embeddings extracted in the Neutral Scenario contain some sensitive information. By using the gender blinded bios and the face embeddings in which demographic information has been removed, we reduced the amount of latent sensitive information within the agnostic embeddings. This reduction leads us to almost random-choice accuracies in the gender case (i.e. in a binary task, the random choice classifier's accuracy is $50$\%), but in the ethnicity one the classifiers fall far from this limit (i.e. $33\%$ corresponding to $3$ ethnic groups), since there is still some information related to that sensitive attribute in the candidate competencies.

\begin{table}[t]
    \centering
    \begin{tabular}{l|c|c|c|c|c|c}
    \hline
    \multirow{2}{*}{\textbf{Scenario}}&\multicolumn{3}{c|}{\textbf{Gender Classification}}&\multicolumn{3}{c}{\textbf{Ethnicity Classification}}\\
    \cline{2-7}
    &\textbf{SVM}&\textbf{RF}&\textbf{NN}&\textbf{SVM}&\textbf{RF}&\textbf{NN}\\
    \hline
    \hline
    Neutral&$65.04\%$&$62.25\%$&$63.92\%$&$54.13\%$&$51.94\%$&$50.29\%$\\
    \hline
    Biased&$89.50\%$&$88.46\%$&$86.37\%$&$76.40\%$&$75.88\%$&$74.31\%$\\
    \hline
    Agnostic&$54.13\%$&$51.94\%$&$52.94\%$&$48.85\%$&$48.13\%$&$49.71\%$\\
    \hline
    \end{tabular}
    
    \caption{Accuracy of different classification algorithms, trained with feature embeddings extracted by the recruitment tool in each scenario (SVM $=$ Support Vector Machines, RF $=$ Random Forests, NN $=$ Neural Networks).}
    \label{tab:demographics}
\end{table}

\section{Conclusions}\label{sec:conclusions}

The development of Human-Centric Artificial Intelligence applications will be critical to ensure the correct deployment of AI technologies in our society. In this paper we have revised the recent advances in this field, with particular attention to available databases proposed by the research community. We have also presented FairCVtest, a new experimental framework (publicly available\footnote{https://github.com/BiDAlab/FairCVtest}) on AI-based automated recruitment to study how multimodal machine learning is affected by biases present in the training data. Using FairCVtest, we have studied the capacity of common deep learning algorithms to expose and exploit sensitive information from commonly used structured and unstructured data.

The contributed experimental framework includes FairCVdb, a large set of $24$,$000$ synthetic profiles with information typically found in job applicants' resumes from different data domains (e.g. face images, text data and structured data). These profiles were scored introducing gender and ethnicity biases, which resulted in gender and ethnicity discrimination in the learned models targeted to generate candidate scores for hiring purposes. In this scenario, the system was able to extract demographic information from the input data, and learn its relation with the biases introduced in the scores. This behavior is not limited to the case studied, where the bias lies in the target function. Feature selection or unbalanced data can also become sources of biases. This last case is common when datasets are collected from historical sources that fail to represent the diversity of our society. 

We discussed recent methods to prevent undesired effects of algorithmic biases, as well as the most widely used databases in the bias and fairness research in AI. We then experimented with one of these methods, known as SensitiveNets, to improve fairness in this AI-based recruitment framework. Our agnostic setup removes sensitive information from text data at the input level, and apply SensitiveNets to remove it from the face images during the learning process. After the demographic ``blinding'' process, the recruitment system did not show discriminatory treatment even in the presence of biases in training data, thus improving equity among different demographic groups.


The most common approach to analyze algorithmic discrimination is through group-based bias \cite{serna2020formulation}. However, recent works are now starting to investigate biased effects in AI with user-specific methods, e.g. \cite{join_optimization}\cite{pentland2020fair}. We plan to update FairCVtest with such user-specific biases in addition to the considered group-based bias. Other future work includes extending our testbed to other multimodal setups like smartphone-based interaction with application to authentication~\cite{2020_CDS_HCIsmart_Acien}, behavior understanding~\cite{2020_AAAI_BeCAPTCHA_Acien}, and remote monitoring/assessment~\cite{2020_AAAI_edBB_JH}. Finally, we also foresee worthy research in the extension of the presented bias-assessment~\cite{2022_SafeAI_IFBiD_Serna} and bias-reduction methods~\cite{SensitiveNets} based on recent advances in biometric template protection~\cite{2017_PR_multiBtpHE_marta} and distributed privacy preservation~\cite{2022_Access_DP-CL_Ahmad}.

\bmhead{Acknowledgments}

This work has received funding from different projects, including BBforTAI (PID$2021$-$127641$OB-I$00$ MICINN/FEDER), HumanCAIC (TED$2021$-$131787$B-I$00$), TRESPASS-ETN (MSCA-ITN-$2019$-$860813$), and PRIMA (MSCA-ITN-$2019$-$860315$). The work of A. \sur{Peña} is supported by a FPU Fellowship (FPU$21$/$00535$) by the Spanish MIU. Also, I. \sur{Serna} is supported by a FPI Fellowship from the UAM.

\section{Compliance with Ethical Standards}\label{sec:ethical_delcaration}

\bmhead{Conflict of Interest}
On behalf of all authors, the corresponding author states that there is no confict of interest.
\bmhead{Ethical Approval}
This article does not contain any studies with human participants performed by any of the authors.
\bmhead{Funding}
This work has received funding from different projects, including BBforTAI (PID$2021$-$127641$OB-I$00$ MICINN/FEDER), HumanCAIC (TED$2021$-$131787$B-I$00$), TRESPASS-ETN (MSCA-ITN-$2019$-$860813$), and PRIMA (MSCA-ITN-$2019$-$860315$). The work of A. \sur{Peña} is supported by a FPU Fellowship (FPU$21$/$00535$) by the Spanish MIU. Also, I. \sur{Serna} is supported by a FPI Fellowship from the UAM.


\bibliography{egbib.bib}


\end{document}